\documentclass[runningheads]{llncs}
\usepackage{graphicx}
\usepackage{times}
\usepackage{latexsym}
\usepackage{url}
\usepackage{graphicx}
\usepackage{enumerate}
\usepackage{booktabs}
\usepackage{multirow}
\usepackage{amsmath}
\usepackage{bm}
\usepackage{multirow}
\usepackage{amsfonts}
\usepackage{algorithm}
\usepackage{algorithmic}

\usepackage[encapsulated]{CJK}
\usepackage{capt-of}

\usepackage{arydshln}

\usepackage{CJKutf8}
\usepackage{color}
\usepackage{xcolor}
\usepackage{tcolorbox}

\definecolor{a1c}{RGB}{242, 135, 47}
\definecolor{a2c}{RGB}{154, 202, 116}
\definecolor{a3c}{RGB}{143, 220, 177}

\begin{document}
\title{Aggressive Language Detection with Joint Text Normalization via Adversarial Multi-task Learning}

\author{Shengqiong Wu \and
Hao Fei \and
Donghong Ji\thanks{Corresponding author.}
}

\institute{Key Laboratory of Aerospace Information Security and Trusted
Computing, Ministry of Education, School of Cyber Science and
Engineering, Wuhan University,Wuhan, China
\email{\{whuwsq,hao.fei,dhji\}@whu.edu.cn} 
}

\maketitle

\begin{abstract}
Aggressive language detection (ALD), detecting the abusive and offensive language in texts, is one of the crucial applications in NLP community.
Most existing works treat ALD as regular classification with neural models, while ignoring the inherent conflicts of social media text that they are quite unnormalized and irregular.
In this work, we target improving the ALD by jointly performing text normalization (TN), via an adversarial multi-task learning framework.
The private encoders for ALD and TN focus on the task-specific features retrieving, respectively, and the shared encoder learns the underlying common features over two tasks.
During adversarial training, a task discriminator distinguishes the separate learning of ALD or TN.
Experimental results on four ALD datasets show that our model outperforms all baselines under differing settings by large margins, demonstrating the necessity of joint learning the TN with ALD.
Further analysis is conducted for a better understanding of our method.
\end{abstract}

\section{Introduction}

Aggressive language detection (ALD) which aims to automatically detect abusive, offensive language and hate speech in social media texts, as one of the important applications of Natural Language Processing (NLP), has recently received increasing research attention.
Yet there are still limited efforts paid for ALD task.
Current works mostly treat ALD as a regular text classification by neural networks, e.g., Long-short Term Memory (LSTM) \cite{hochreiter1997long}, Convolutional Neural Networks (CNN) \cite{kim2014convolutional} or Transformer \cite{vaswani2017attention}, with sophisticated features, e.g., pre-trained embeddings \cite{badjatiya2017deep,zhang2018detecting}.

\begin{figure}
\includegraphics[width=1.0\textwidth]{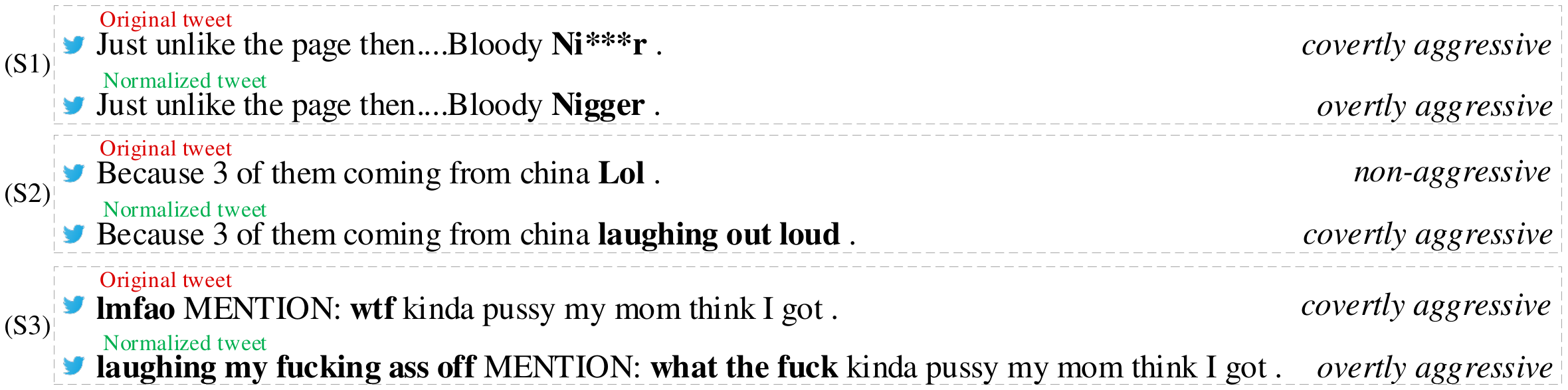}
\caption{
Example tweets for illustrating the aggressive language detection task under unnormalized and normalized contents, respectively.
On the right side of the sentences are the corresponding labels predicted by a detector.
The labels for \textcolor{a3c}{\emph{normalized tweets}} are correct as ground truth.
}
\label{tweet examples}
\end{figure}

Nevertheless, social media texts often differ substantially from the written texts, that is, social media texts can be much noisy and contain typos \cite{hassan-menezes-2013-social,yang-eisenstein-2013-log}, e.g., abbreviations, letter repetition, etc.
Such characteristic of unnormalized texts can greatly hinder the detection of aggressive contents.
Taking the examples sentence (S1-S3) in Fig. \ref{tweet examples}, the raw unnormalized expressions that carry crucial signals for indicating offensive languages, can be difficult for a detector to give correct prediction when merely seeing the surface forms.
However, if these unnormalized contents are transformed into the normalized standard texts, the inferences of the detector can be much easier.

\begin{figure}[!t]
\centering
\includegraphics[width=0.94\textwidth]{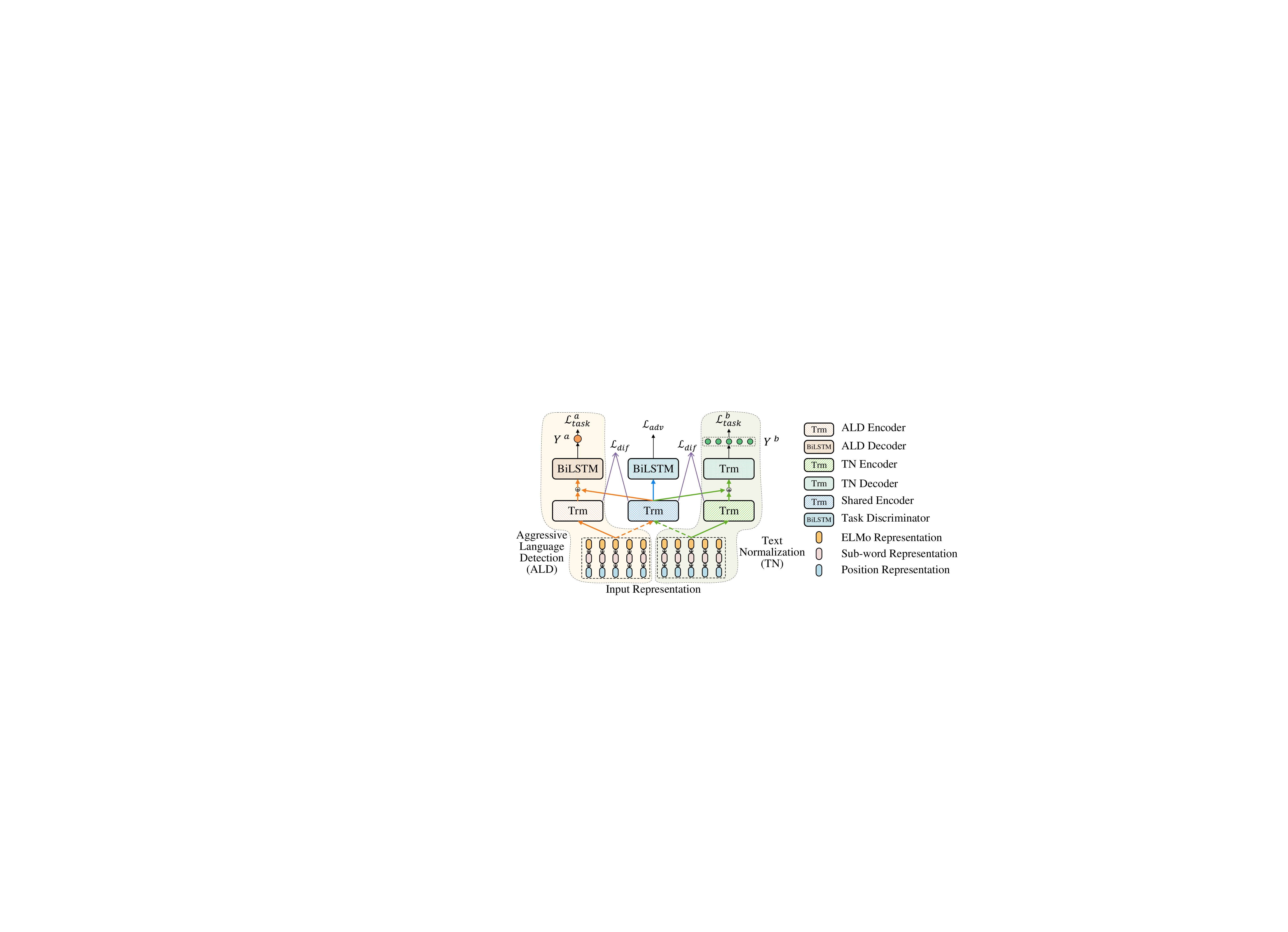}
\caption{
The overall framework.
During adversarial training, the ALD\&shared part will perform adversarial propagation (i.e., \textcolor{a1c}{$\longrightarrow$}\&\textcolor{a1c}{$\dashrightarrow$}), taking turn with the TN\&shared part (i.e., \textcolor{a2c}{$\longrightarrow$}\&\textcolor{a2c}{$\dashrightarrow$}).
}
\label{overall framework}
\end{figure}

Based on the above observation, in this paper, we propose to improve the ALD task by simultaneously handling the text normalization (TN).
A multi-task learning (MTL) framework is adopted for the joint training of these two tasks.
As depicted in Fig. \ref{overall framework}, first, the shared encoder is expected to learn the underlying common features over two tasks, while the private encoders for ALD and TN learn the task-relevant features, respectively, based on which the decoders can make their own task predictions.
To further enhance the capabilities of the shared and private feature representations, respectively, we suggest the adversarial training architecture \cite{liu-etal-2017-adversarial}.
Technically, a task discriminator is used for distinguishing the separate learning of ALD and TN tasks.

We conduct experiments on four widely used ALD datasets, including TRAC \cite{kumar2018benchmarking}, HSOL \cite{davidson2017automated}, KTC \cite{brassard2019subversive} and OLI \cite{zampieri2019semeval},
based the annotated text normalization data, Lexnorm15 \cite{baldwin-etal-2015-shared}. 
Results show that the aggressive language detection can benefit much from the joint learning with text normalization.
Our model outperforms baseline methods by a large margin, with 64.0\% and 53.6\% F1 score in TRAC-FB and TRAC-TW test sets, respectively, and average 90.5\% F1 score for other three datasets.
In-depth analysis is performed for further understanding of how the TN influences the ALD task, as well as the mechanism of our proposed adversarial multi-task learning framework.

\section{Related Work}
\label{Related Works}

Aggressive language detection (ALD) has received increasing research attention in NLP community.
ALD is traditionally tackled as a regular text classification task, which is often approached with types of surface features such as token frequencies, text characteristic, linguistic features, and word embeddings  \cite{badjatiya2017deep,schmidt2017survey,zhang2018detecting}.
Initial works employ the statistical machine learning algorithms for the tasks \cite{schmidt2017survey,wulczyn2017ex}.
More recently, neural networks, e.g., LSTM, CNN and Transformer, etc. are extensively adopted as de-facto methods for yielding state-of-the-art task performances \cite{Fei9113297,FEI2020102311} and capture semantics of texts \cite{fei-etal-2020-cross,FeiZRJ20}.
For example, some researches use the CNN as encoder to capture the n-gram features in the texts \cite{gamback2017using,gao2018hierarchical}.
Nikhil et al. (2018) exploit the LSTM model with an attention unit, which is efficient on constructing sentence representations \cite{nikhil2018lstms}.
Further, Zhang et al. (2018) use a combination of CNN and gated recurrent unit (GRU) for detecting the hate speech on twitter texts\cite{zhang2018detecting}.
In this work, we consider improving the ALD task by simultaneously performing the text normalization, as the social media texts often involve much noisy and unnormalized expressions.

Our work also relates closely to the application of multi-task learning (MTL) technique.
MTL provides an avenue for effectively integrating multiple standalone single tasks into shared one, which has been extensively exploited to a wide range of NLP tasks for achieving improved performances \cite{FEI2020241,liu2016recurrent}.
There are several works utilizing MTL framework for ALD task \cite{cimino2018multi,vaidya2019empirical}.
For example, Cimino et al. (2018) employ a shared
Bi-LSTM to exploit the related information between the labels. 
Vaidya et al. (2019) propose a multi-task learning model that jointly learns to predict the toxicity of a comment as well as the identities present in the comments.
Different from these methods, we propose to conduct joint learning for text normalization and classification via MTL in soft paramter sharing with a shared-private structure.
Besides, we equip our MTL framework with the adversarial training algorithm, which is also a crucial technique for building stronger MTL models and bringing improvements \cite{liu-etal-2017-adversarial}.

\section{Framework}
\label{Method}

As shown in Fig.\ref{overall framework}, our multi-task learning framework makes prediction for two shared tasks by taking as inputs two types of sources, respectively.
The aggressive language detection task $T^a$ is modeled as sentence-level classification, predicting class labels $Y^a$ based on input sentence ${X}^{a}=\{x^{a}_{1},\cdots,x^{a}_{n^{a}}\}$.
And the text normalization task $T^b$ is formulized as a sequence generation task, to yield the normalized word sequence $Y^a=\{y^{b}_{1},\cdots,y^{b}_{m^{b}}\}$ from the input unnormalized sentence ${X}^{b}=\{x^{b}_{1},\cdots,x^{b}_{n^{b}}\}$.

The overall architecture is based on shared-private multi-task structure, mainly consisting of four tiers: input representation, shared and private encoders, task-specific decoders and task discriminator. 
We first embed the inputs for $T^a$ or $T^b$ into vectorial representations, respectively.
Then, the shared encoder and the private encoder for the task $T^a$ or $T^b$ learn the feature representation by taking the input representations, respectively.
Based on the shared and the private feature representations, the decoder for task $T^a$ or task $T^b$ finally make their own predictions, respectively.
We note that the part for $T^a$ and the part for $T^b$ takes turn to perform the learning once at a time.
During the shifting of the learning for task $T^a$ or $T^b$, the task discriminator based on the shared feature representation predicts the task-id label $Y^d$.

\subsection{Input Representation}

Note that the input representations for aggressive language detection and for text normalization are constructed with same manner, and thus for brevity, we do not distinguish this in the notations with superscript e.g., $x^a,x^b$.
We mainly consider three types of input features, including the surface word representation, the sub-word representation and the position representation.
For each word $x_t$, we enhance the representation capability by employing the contextualized language model, ELMo \cite{peters2018deep}, as $\bm{v}^{ELMo}_t$.
Sub-word level word representations have been shown useful to relieve the noises in unnormalized texts \cite{lal2019mixing}.
We thus use a character-level CNN to generate sub-word embeddings for each word, denoted as $\bm{v}^{sbw}_t$.
We then consider capturing the order information about the relative or absolute position of the tokens.
Concretely, we use a lookup table to obtain the position embedding $\bm{v}^{pe}_t$ for each input word.
We finally concatenate these representations into unified input representation: $\bm{x}_t = [\bm{v}^{ELMo}_t;\bm{v}^{sbw}_t;\bm{v}^{pe}_t]$

\subsection{Shared and Private Feature Encoder}

We consider the self-attention based Transformer (Trm) model \cite{vaswani2017attention} as our major shared and private encoders, due to its prominence on feature mining \cite{FeiRJ19,liu2016recurrent}.
Technically, in Transformer encoder, the input $\bm{x}$ is first mapped into queries $\bm{Q}$, values $\bm{V}$, and keys $\bm{K}$ via linear projection.
We then compute the relatedness between the $\bm{K}$ and $\bm{Q}$ via Scaled Dot-Product alignment function, which is multipled by $\bm{V}$:
\begin{equation}
\setlength\abovedisplayskip{3pt}
\setlength\belowdisplayskip{3pt}
    \label{self attention}
    \bm{\alpha} = \text{softmax}(\frac{\bm{Q}\cdot\bm{K}^{\mathrm{T}}}{\sqrt{d_{k}}}) \cdot \bm{V}
\end{equation}
where $d_{k}$ is a scaling factor.
$h$ parallel attention heads can focus on different parts of channels of the value vectors.
Finally, all the vectors produced by parallel heads are concatenated together to form a unified representation.
\begin{equation}
\setlength\abovedisplayskip{3pt}
\setlength\belowdisplayskip{3pt}
    \label{multi-head}
    \bm{R} = [\bm{\alpha}_{1}; \cdots; \bm{\alpha}_{h}] \cdot \bm{W}_{\alpha} + b_{\alpha} \,.
\end{equation}
We can summarize all the above calculations of the Transformer encoder as follows:
\begin{equation}
\setlength\abovedisplayskip{3pt}
\setlength\belowdisplayskip{3pt}
    \label{encoder}
    \bm{r}_1,\cdots,\bm{r}_n = \text{Trm}(\bm{x}_1,\cdots,\bm{x}_n)
\end{equation}

We employ three multi-layer Transformer encoders for the feature learning of 
the task $T^a$, the task $T^b$ and the shared common one, respectively, which are denoted as:
\begin{equation}
\setlength\abovedisplayskip{3pt}
\setlength\belowdisplayskip{3pt}
\begin{aligned}
    \bm{r}^a_1,\cdots,\bm{r}^a_{n^a} &= \text{Trm}^{a}(\bm{x}^a_1,\cdots,\bm{x}^a_{n^a}) \\
    \bm{r}^b_1,\cdots,\bm{r}^a_{n^b} &= \text{Trm}^{b}(\bm{x}^b_1,\cdots,\bm{x}^b_{n^b}) \\
    \bm{r}^{s}_1,\cdots,\bm{r}^s_{n^{*}} &= \text{Trm}^{s}(\bm{x}^{*}_1,\cdots,\bm{x}^{*}_{n^{*}})
\end{aligned}
\end{equation}
where $\bm{x}^{*}$ indicates that the input of the shared encoder can either be the source of text normalization, or the one of aggressive language detection.

\subsection{Task-specific Decoders}

\paragraph{\bf Aggressive language detector.}
We use the BiLSTM as the decoder for aggressive language detection.
Given the shared and private feature representation, $\bm{r}^s$ and $\bm{r}^a$, the softmax is expected to predict the resulting label:
\begin{equation}
\setlength\abovedisplayskip{3pt}
\setlength\belowdisplayskip{3pt}
\begin{aligned}
    \label{class prediction}
    \bm{h}^a &= \text{BiLSTM}(\hat{\bm{r}}^a_1,\cdots,\hat{\bm{r}}^a_{n^a}) \\
    Y^b &= \text{softmax}(\bm{h}^a) \,.
\end{aligned}
\end{equation}
where $\hat{\bm{r}}^a_t$ is the concatenation of the shared and the corresponding private feature representations, i.e., $[\bm{r}^s_t;\bm{r}^a_t$].

\paragraph{\bf Text normalizer.}
Given an input unnormalized sentence, the goal of the text normalization is to generate the normalized word sequence counterpart.
We consider it as text generation task, via a asynchronous sequence-to-sequence scheme \cite{dehghani2018universal}.
We use the same architecture of the decoder in Vaswani et al. (2017) \cite{vaswani2017attention} for neural machine translation, which is also a multi-layer Transformer module with element-wise softmax classifier.
Technically, the decoding can be described as:
\begin{equation}
\setlength\abovedisplayskip{3pt}
\setlength\belowdisplayskip{3pt}
\begin{aligned}
    \bm{h}^b_1,\cdots,\bm{h}^b_{m^b} &= \text{Trm}^{b}(\hat{\bm{r}}^b_1,\cdots,\hat{\bm{r}}^b_{n^b}) \\
    y^b_1,\cdots,y^b_{m^b} &= \text{softmax}^{b}(\bm{h}^b_1,\cdots,\bm{h}^b_{m^b}) \\
\end{aligned}
\end{equation}
where $\hat{\bm{r}}^b_t$ is the concatenation of the output representations of the shared and the corresponding private encoder.
We finally can obtain the normalized texts $y^b_1,\cdots,y^b_{m^b}$.

\subsection{Task Discriminator}
Although the shared and private encoders separate the feature space into the shared and private one, there are still chances that the learnt features for text normalization and aggressive language detection are entangled, disturbing the encoders to focus on their own roles.
Therefore, we employ a third-party task discriminator with adversarial training to refine the features \cite{liu-etal-2017-adversarial,zhou-etal-2019-dual}.
Our discriminator $\mathcal{D}$ is a binary classifier for predicting what is the current task, based merely on the shared feature representation $\bm{r}^s$.
Ideally, once the discriminator cannot accurately identify the task, the shared feature can be understood as most purified one.
Specifically, we use a BiLSTM with softmax: 
\begin{equation}
\setlength\abovedisplayskip{3pt}
\setlength\belowdisplayskip{3pt}
\begin{aligned} \label{discriminator}
    \bm{h}^s &= \text{BiLSTM}(\hat{\bm{r}}^s_1,\cdots,\hat{\bm{r}}^s_{n^{*}}) \\
    Y^d &= \text{softmax}(\bm{h}^s) \,.
\end{aligned}
\end{equation}
where $Y^d$ is the task id for representing the current task, i.e., ALD or TN.

\subsection{Learning}

For each task $T^a$ or $T^b$, we minimize the cross-entropy of the predicted and gold one:
\begin{equation}
\setlength\abovedisplayskip{3pt}
\setlength\belowdisplayskip{3pt}
    \mathcal{L}_{task} = - [ \begin{matrix} \sum_{j=1}^{N^a} \end{matrix} \hat{Y}^{a}_j \log(Y^{a}_{j}) + \begin{matrix} \sum_{j=1}^{N^b} \end{matrix} \hat{Y}^{b}_j \log(Y^{b}_{j}) ]
\end{equation}
where $\hat{Y}^{a}$ and $\hat{Y}^{b}$ are the gold annotations for each task, respectively.

The target for adversarial training is to urge the shared features such that the task discriminator cannot reliably predict the task id:
\begin{equation}
\setlength\abovedisplayskip{3pt}
\setlength\belowdisplayskip{3pt}
    \label{adversarial loss}
    \mathcal{L}_{adv} = \mathop {\min }\limits_{\theta_{S}}( \mathop{\max} \limits_{ \mathcal{D}} ( \begin{matrix} \sum_{j=1}^{N^a} \end{matrix} \hat{Y}^{d_a} \log(Y^{d_a})  +  \begin{matrix} \sum_{j=1}^{N^b} \end{matrix} \hat{Y}^{d_b} \log(Y^{d_b})   ))
\end{equation}
where $\hat{Y}^{d_a}$ and $\hat{Y}^{d_b}$ represent each ground-truth task id, respectively.
In addition, we impose an orthogonality constraint for further penalizing redundant latent representations between the shared and private features:
\begin{equation}
\setlength\abovedisplayskip{3pt}
\setlength\belowdisplayskip{3pt}
    \label{diff loss}
    \mathcal{L}_{dif} = ||\bm{r}_a^T \bm{r}_s||_{F}^{2} + ||\bm{r}_b^T \bm{r}_s||_{F}^{2}
\end{equation}
where $||\cdot||_{F}^{2}$ is the squared Frobenius norm. 
The final loss of the overall framework is:
\begin{equation}
    \label{loss}
    \mathcal{L} = \mathcal{L}_{task} + \lambda\mathcal{L}_{adv} + \beta\mathcal{L}_{dif}
\end{equation}
where $\lambda$ and $\beta$ are two coupling co-efficiency for regulating the learning.

\paragraph{\bf Training details.}
During adversarial training, the private encoders for TN and ALD task will take turn to perform forward propagation along with the shared encoder, within the multi-task framework as depicted in Fig.\ref{overall framework}.
Besides, we find in our preliminary experiment that directly training the whole framework with cold-start can be difficult and causes high variance.
Thus we consider the warm-start method, that is, we first pre-trained the TN part, and until it is close to the convergence we then jointly train the ALD module.
During each turn of the TN or ALD training, the shared feature encoder and task discriminator will be trained to reach an agreement, at which both of them do not improve, and the discriminator fail to differentiate among both the languages.
We keep such training iterations until the overall performance reaches its plateau.

\section{Experiments}
\label{Experiments}

\subsection{Settings}

\begin{table}[!t]
\begin{center}
\caption{Statistics of the five datasets.}
\label{statistics of dataset}
\begin{tabular}{lc|ccc|c}
\hline
\quad \bf Dataset  &  \bf Source
 &  \bf Train & \bf Develop &  \bf Test  & \bf Label\\
\hline
\quad TRAC & Facebook, Twitter & 12,000 & 3,000 &  916(FB)/1,257(TW) & 3  \\
\quad  HSOL & Twitter  & 22,304 & - & 2,479  & 3 \\
\quad  KTC & Wikipedia & 143,613 & 15,917 & 63,678  &  2 \\
 \quad OLI & Twitter & 11,915 & 1,325 & 860  &  2 \\
\hdashline
 \quad Lexnorm15 & Twitter & 2,875 & - & 2,024  & - \\
\hline
\end{tabular}
\end{center}
\end{table}

\paragraph{\bf Dataset.}

We evaluate our model mainly on four English datasets, as follows.
1) The \texttt{TRAC} dataset is published in a shared task\footnote{\url{https://sites.google.com/view/trac1/home}} for ALD.
The sources are from English social media, e.g., Facebook and Twitter, and there are two corresponding testing sets, i.e., \texttt{FB} and \texttt{TW}.
There are three labels for indicating the aggression degree: covertly aggressive(CAG), non-aggressive(NAG) and overtly aggressive(OAG).
The other three datasets are also widely used for hate speech or offensive language detection, including 
2) \texttt{HSOL} (Hate Speech and Offensive Language)\footnote{\url{https://github.com/t-davidson/hate-speech-and-offensive-language}}, 
3) \texttt{OLI} (Offensive Language Identification)\footnote{\url{https://competitions.codalab.org/competitions/20011}}
4) \texttt{KTC} (Kaggle Toxicity Competition)\footnote{\url{https://www.kaggle.com/c/jigsaw-toxic-comment-classification-challenge}}.
Besides, we employ the labeled text normalization dataset \texttt{Lexnorm15}\footnote{\url{https://noisy-text.github.io/2015/index.html}}, where each unnormalized sentence has a normalized counterpart sentence as annotation supervision.
In Table \ref{statistics of dataset} we show the detailed statistics of each dataset.

\begin{table*}[!t]
\begin{center}
  \caption{
\label{main table}
  Main results for ALD on \texttt{TRAC} dataset and TN on \texttt{Lexnorm15} dataset.
Results with $\ddagger$ indicate that the additional resources are used.
`w/o ELMo' indicate replacing the ELMo representations with randomly initialized ones.
  }
  \begin{tabular}{l ccc ccc c}
\toprule
\multirow{2}{*}{} & \multicolumn{3}{c}{\texttt{FB}}  & \multicolumn{3}{c}{ \texttt{TW} } & \multicolumn{1}{c}{\texttt{Lexnorm15}}  \\ 
\cmidrule(r){2-4}\cmidrule(r){5-7}\cmidrule(r){8-8}
&  Precision &  Recall &  F1 &  Precision &  Recall &  F1  &  F1 \\
\midrule 
\multicolumn{8}{l}{$\bullet$  \bf ALD (Standalone) }\\
\quad RCNN & 71.9 & 48.9 & 53.6 & 49.5& 52.6 & 46.2 & - \\
\quad CNN+GRU & 66.7 & 55.0 & 58.6 & 46.6 & 47.8 & 46.3 & - \\
\quad Transformer& 56.3 & 62.7 & 59.0 & 50.0  & 48.9 & 47.6 & - \\
\hdashline
\quad Ramiandrisoa et al. (2018) \cite{ramiandrisoa2018irit}$\ddagger$ & - & - & 57.6 & - & - & 51.1  & - \\
\quad Madisetty et al. (2018) \cite{madisetty2018aggression}$\ddagger$ & - & - & 60.4 & - & - & 50.8  & - \\
\hline
\multicolumn{7}{l}{$\bullet$  \bf ALD (with Pipeline TN)} & 46.1\\
\quad RCNN & 67.3 & 50.8 & 54.8 & 47.4 & 48.5 & 47.3 &  \\
\quad CNN+GRU & 68.4 & 57.3 & 60.4 & 50.3 & 51.9 & 48.6 &  \\
\quad Transformer& 63.6 & 59.1 & 60.8 & 51.5  & 52.8 & 49.8 &  \\
\hline\hline
\multicolumn{8}{l}{$\bullet$  \bf  ALD (with Joint TN)}\\
\quad GRU & 68.7 & 53.3 & 57.6 & 51.5 & 53.0 & 50.5   & 45.4 \\
\quad Transformer & 69.7 & 59.1 & 61.9 & 53.3 & 52.4 & 51.9  &  47.2 \\
\hdashline
\quad Ours & 70.7 & 60.8 & \textbf{64.0} & 54.6 & 53.9 & \textbf{53.6}  &  \bf48.2 \\
\quad \quad w/o ELMo & 69.5 & 58.7 & 62.1 & 53.3 & 54.9 & 52.5 & 46.0 \\
\bottomrule
  \end{tabular}
\end{center}
\end{table*}

\paragraph{\bf Baselines and evaluation.}
We mainly compare our model with baseline methods under three types of settings.
\textbf{1)} In the first setting, we show the performances of the standalone ALD. 
We make comparisons with the RCNN model \cite{lai2015recurrent}, CNNs+GRU model \cite{zhang2018detecting}, and Transformer.
Also we show the current state-of-the-art results by Madisetty et al. (2018) \cite{madisetty2018aggression} and Ramiandrisoa et al. (2018) \cite{ramiandrisoa2018irit}.
\textbf{2)} In the second setting, we evaluate the performances of ALD where the raw unnormalized sentences are offline pre-processed into the normalized ones by a well pre-trained TN model.
\textbf{3)} We lastly compare the performances by the joint learning of ALD and TN.
We compare with the models e.g., GRU and Transformer under MTL.
We adopt the standard Precision, Recall, and use the weighted F1 score\footnote{
A variant of macro F-score that takes into consideration the instance numbers for each label.
It can result in a value that is not between precision and recall.
} as the metrics, following existing work \cite{ramiandrisoa2018irit}.

\paragraph{\bf Pre-processing and hyperparameter.}

We use the ELMo\footnote{\url{https://allennlp.org/elmo}} to offer the default word representations.
Besides, we also employ the pre-trained Glove\footnote{\url{https://nlp.stanford.edu/projects/glove/}} and the BERT\footnote{\url{https://github.com/google-research/bert}, base-cased-version.}.
For ALD/TN encoder and TN decoder, we use the 2-/3-/3-layer version of Transformer, respectively.
For ALD decoder and task discriminator, we use the 1-/2-layer version of BiLSTM.
the pre-trained model ELMo in which LSTM hidden size is 1024 and the output size is 128. 
We set mini-batch size as 32 for TN, and 16 for ALD.
To avoid overfitting, we adjust the dropout rate to 0.4.
Considering the number of training data for TN and ALD are imbalanced, we get more TN data by using a slang dictionary\footnote{\url{https://github.com/cbaziotis/ekphrasis}} to correct certain typos in ALD dataset.
We use Adam as the optimizer with early-stop strategy.
For each task, we take the hyperparameters which achieve the best performance on the development set via a small grid search over combinations of the initial learning rate $[0.001, 0.0001]$, $\lambda \in [0.01, 0.1]$, $\beta \in [0.01, 0.1]$. 
Finally, we chose the learning rate as 5e-4, $\lambda$ as 0.05, and $\beta$ as 0.01.
We open our model implementation at \url{https://github.com/ChocoWu/ALD-TN}.

\begin{table}[!t]
\begin{minipage}{\textwidth}
\centering
\begin{minipage}[b]{0.54\textwidth}
\setlength{\tabcolsep}{2.7pt}
\captionof{table}{
Ablation results (F1 scores).
    }
\label{ablation}
\centering
\begin{tabular}{lccc}
\toprule
& \multicolumn{1}{c}{\texttt{FB}}  & \multicolumn{1}{c}{ \texttt{TW} } & \multicolumn{1}{c}{\texttt{Lexnorm15}}  \\ 
\midrule 
 Ours & {64.0} & {53.6} & 48.2 \\
\quad by \emph{cold-start} &  59.0 &  52.2 &  40.1 \\
\hline
\multicolumn{4}{l}{$\bullet$  \bf Shared-private with Adversarial training}\\
\quad w/o $\mathcal{L}_{adv}$ &  63.6 &  51.5 &  46.0 \\
\quad w/o $\mathcal{L}_{dif}$ & 62.9 &  50.8 &  45.8 \\
\quad w/o $\mathcal{L}_{adv}$\&$\mathcal{L}_{dif}$ &  62.3 &  49.9 & 45.4 \\
\hline
\multicolumn{4}{l}{$\bullet$  \bf Pre-trained word representation}\\
\quad +Glove &  62.5 &  46.6 &  45.7 \\
\quad +BERT &  \bf 65.2& \bf 55.5 & \bf 50.2 \\
\bottomrule
  \end{tabular}
\end{minipage}
\hfill
\begin{minipage}[b]{0.44\textwidth}
\centering
\setlength{\tabcolsep}{2pt}
\captionof{table}{
Results (F1 scores) on the other datasets.
}
\label{other datasets results}
 \begin{tabular}{lcccc}
\toprule
 & \multicolumn{1}{c}{\texttt{OLI}}  & \multicolumn{1}{c}{ \texttt{KTL} } & \multicolumn{1}{c}{\texttt{HSOL}} & \texttt{Avg.} \\ 
\midrule 
\multicolumn{5}{l}{$\bullet$  \bf  Standalone}\\
\quad CNN+GRU &  76.1 &  90.0 &  89.2 & 85.1 \\
\quad Transformer&  80.1 &  91.4 &  88.1 & 86.5 \\
\hline
\multicolumn{5}{l}{$\bullet$  \bf  with Joint TN}\\
\quad GRU &  80.7 &  92.0 &  88.9 & 87.2 \\
\quad Transformer&  81.2 &  92.1 &  90.6 & 87.9 \\
\hdashline
\quad Ours  &  \bf 83.7 & \bf93.4 & \bf94.4  & \bf 90.5 \\
\hline
  \end{tabular}
\end{minipage}
\end{minipage}
\end{table}

\subsection{Main Results}

In Table \ref{main table} we summarize the main results.
The first observation we can notice is that the ALD with normalized texts are universally better than the standalone ALD, which demonstrates the necessity of the text normalization for the ALD texts.
Also the existing works with additional resources, e.g., pre-trained embeddings and sentimental lexicons, can greatly improve the ALD performances.
Second, the ALD jointly training with TN can perform better than that with pipeline TN.
For example the ALD performances by Transformer model in joint TN setting obtains 61.9\% and 51.9\% F1 scores, being better than that in the pipeline TN with 60.8\% and 49.8\% F1 scores.
The underlying possible reason largely lies in that the joint learning of two tasks can avoid introducing noises from TN to ALD.
In addition, such joint training also mutually benefits the text normalization task, as can be seen by the TN results with 46.1\% F1 score in pipeline ALD and 48.2\% F1 score in joint ALD (by GRU model), respectively.

Most importantly, our proposed model gives the overall best results than all the baselines by large margins, with 64.0\% and 53.6\% F1 scores on two ALD test sets, and 48.2\% F1 score for TN.
This proves the effectiveness of the proposed method for aggressive language detection.
We further remove the help of ELMo contextualized word representations, and find that
our performances are still stronger than baselines.
We note that the differences between ours and the Transformer model in ALD with joint TN setting are the shared-private structure and the adversarial training with discriminator.
We can see that our results without ELMo keep better than that of the Transformer, verifying the superiority by adopting such enhanced multi-task learning architecture.

\paragraph{\bf Ablation results.}

We conduct ablation study to investigate the contributions of different aspects of our method, as shown in Table \ref{ablation}. 
First of all, we train the framework with warm-start strategy.
When we use the cold-start instead, we find the results got hurt for both two tasks, especially for the text normalization.
We next explore the shared-private structure with adversarial training, by ablating the losses, $\mathcal{L}_{adv}$, $\mathcal{L}_{dif}$, and we find the results will drop, correspondingly.
Notably, without the adversarial part (i.e., without $\mathcal{L}_{adv}$\&$\mathcal{L}_{dif}$), the performances degrade dramatically.
Further, when replacing the default ELMo representation with Glove and BERT pre-trained embeddings, we can receive the corresponding performance decreases and increases, respectively.

\paragraph{\bf Results on the other datasets.}
In Table \ref{other datasets results} we can see that 
the overall trends on the other datasets are similar with that in Table \ref{main table}.
The results by jointly training models with TN are universally stronger than that of the standalone aggressive language detection.
Our model can bring the best results with average 90.5\% F1 score.
This demonstrates the generalization ability of our model on the ALD task.

\begin{figure}[tp]
\begin{minipage}{\textwidth}
\begin{minipage}[b]{0.49\textwidth}
\centering
\includegraphics[scale=.88]{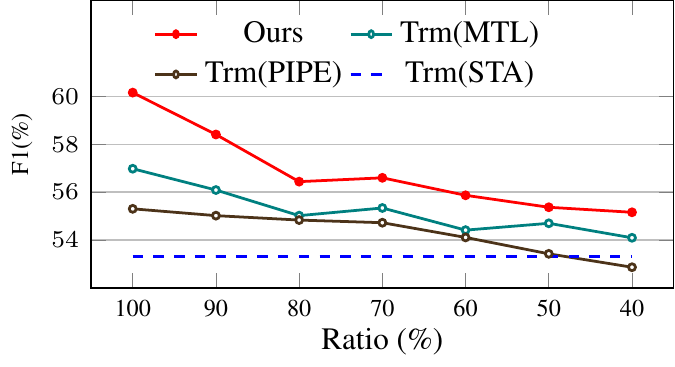}
\captionof{figure}{
ALD performances under a variable ratio of the Lexnorm15 training data.
`MTL' means joint multi-task learning, `PIPE' means pipeline TN, `STA' is standalone ALD.
}
\label{diff-ratio}
\end{minipage}
\hfill
\begin{minipage}[b]{0.49\textwidth}
\centering
\includegraphics[scale=.88]{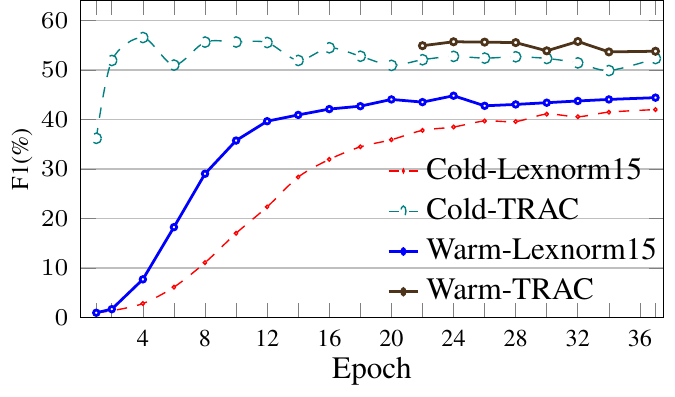}
\captionof{figure}{
Training curves by cold-start and warm-start manner.
Performances of `TRAC' are averaged F1 scores over `FB' and `TW'.
}
\label{diff-train-ways}
\end{minipage}
\end{minipage}
\end{figure}

\subsection{Discussion}

\paragraph{\bf Influences by text normalization training data.}

We introduce the joint training of ALD with TN under adversarial multi-task framework.
We now explore the impacts of ALD task by different numbers of TN training data.
In Fig. \ref{diff-ratio} we show the performances by our model and the Transformer (Trm) model under a variable ratio of training samples of TN.
First, the overall results drop when cutting down the training signals for TN, gradually.
Besides, all the performances under different settings keep a similar trend to the above conclusions, that is, the joint training is more useful than that the pipeline manner, and both superior to the standalone manner.
Notably, our adversarial MTL model is most effective on counteracting the data scarcity of TN, being most robust.

\paragraph{\bf Warm-start training for multi-task framework.}

We study the training effects by using cold-start and warm-start strategy, respectively.
From the patterns in Fig. \ref{diff-train-ways}, clearly, by warm-start training manner, the framework tends to converge to a better results, meanwhile with much more stable learning for both the ALD and TN tasks.
On the contrary, cold-start training of the model introduces turbulences.
This suggests the imperative to use a warm-start training strategy for our adversarial multi-task framework.

\subsection{Case Study}

\begin{figure}[!t]
\includegraphics[width=\textwidth]{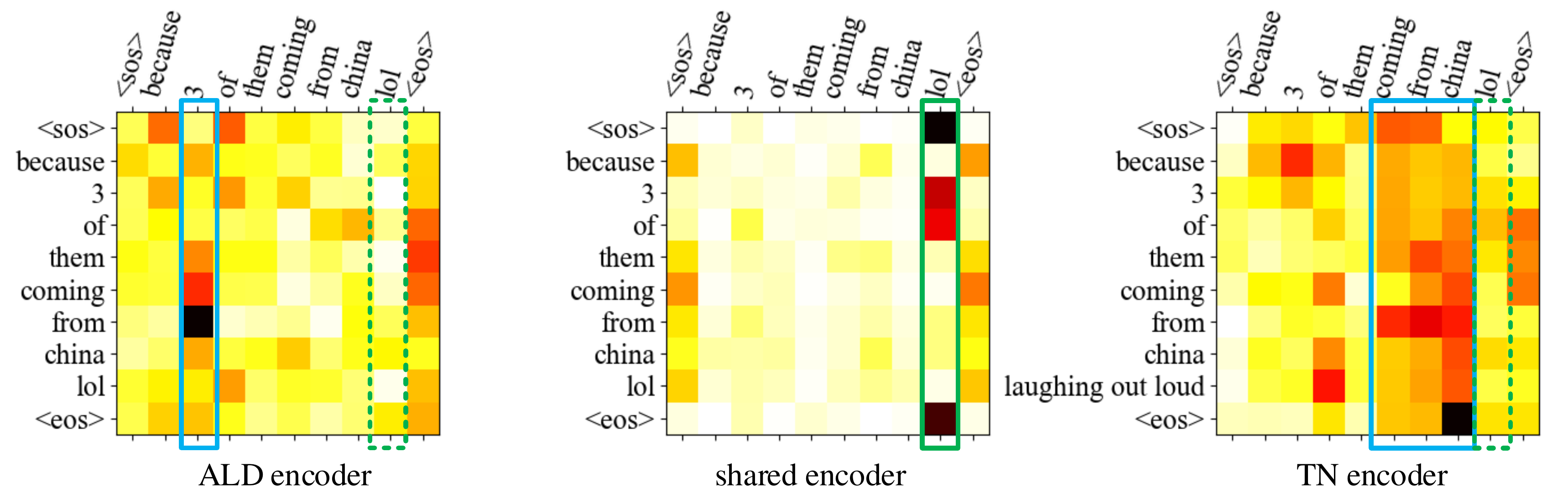}
\caption{
Attention visualizations on ALD encoder, shared encoder and TN encoder, respectively.
On the upper are the input sentences, on the left are the output sentences.
}
\label{attention visual}
\end{figure}

Lastly, we perform case study to see how the task-specific private encoders and the shared encoder under adversarial multi-task training collaborate the learning for ALD.
We empirically visualize the attention on ALD, TN and shared encoders, respectively, based on one correctly inferred example from test set, as can be seen in Fig. \ref{attention visual}.
Interestingly, different encoders can largely focus on their separate roles.
For example, the shared encoder learns the common shared features, paying more attention on the token \emph{`lol'}, which is an kernel clues for both the TN and the ALD.
On the one hand, the \emph{`lol'} corresponding to the normalized phrase \emph{`laughing out loud'} in shared encoder, combined with the relevant clues in TN encoder, are correctly captured by the TN module, leading to a successful prediction.
In the meantime, the \emph{`lol'} also as an important signal captured by shared encoder, together with the other cue features (i.e., \emph{`3'}) by the ALD private encoder, help to result in a correct detection for aggressive language.

\section{Conclusion}
\label{conclusion}
In this work, we proposed to improve the aggressive language detection (ALD) by jointly performing text normalization (TN), via a adversarial multi-task learning framework.
The private encoders for ALD and TN focused on the task-specific feature retrieving, respectively, and the shared encoder learned the underlying common features over two tasks.
During adversarial training, the task discriminator distinguished the separate learning of ALD or TN.
Experimental results on four ALD datasets showed that our model outperformed all baselines by large margins under differing settings, demonstrating the necessity of joint learning the TN with ALD.

\section{Acknowledgments}

This work is supported by the National Natural Science Foundation of China (No.61772378), 
the National Key Research and Development Program of China (No.2017YFC1200500), 
the Humanities-Society Scientific Research Program of Ministry of Education (No.20YJA740062),
the Research Foundation of Ministry of Education of China (No.18JZD015), and 
the Major Projects of the National Social Science Foundation of China (No.11\&ZD189).

\bibliography{refs}
\bibliographystyle{splncs04}

\end{document}